\documentclass[11pt,a4paper]{article}
\usepackage{acl2019}
\usepackage{times}
\usepackage{bbm}
\usepackage{latexsym}
\usepackage{graphicx}
\usepackage{amsmath}
\graphicspath{ {./} }
\usepackage[noend]{algpseudocode}
\usepackage{algorithmicx,algorithm}

\aclfinalcopy 
\def\aclpaperid{301} 

\title{Automated Curriculum Learning for Turn-level Spoken Language Understanding with Weak Supervision}

\author{Hao Lang \\
Alibaba Group \\
\texttt{hao.lang@taobao.com} \\ \And
Wen Wang \\
Alibaba Group (U.S.) Inc. \\
\texttt{w.wang@alibaba-inc.com}}

\date{}

\begin{document}
\maketitle
\begin{abstract}
We propose a learning approach for turn-level spoken language understanding, which facilitates a user to speak one or more utterances compositionally in a turn for completing a task (e.g., voice ordering). A typical pipelined approach for these understanding tasks requires
non-trivial annotation effort for developing its multiple components. Also, the pipeline is difficult to port to a
new domain or scale up. To address these problems, we propose an end-to-end statistical model with weak supervision. We employ randomized beam search with memory augmentation (RBSMA) to solve complicated problems for which long promising trajectories are usually difficult to explore. Furthermore, considering the diversity of problem complexity, we explore automated curriculum learning (CL) for weak supervision to accelerate exploration and learning. We evaluate the proposed approach on real-world user logs of a commercial voice ordering system. Results demonstrate that when trained on a small number of end-to-end
annotated sessions collected with low cost, our model performs comparably to the deployed pipelined system, saving the development labor over an order of magnitude. The RBSMA algorithm improves the test set accuracy by 7.8\% relative compared to the standard beam search. Automated CL leads to better generalization and further improves the test set accuracy by 5\% relative.
\end{abstract}

\section{Introduction}
\label{sect:intro}

\begin{figure}[h]
\includegraphics[scale=0.45]{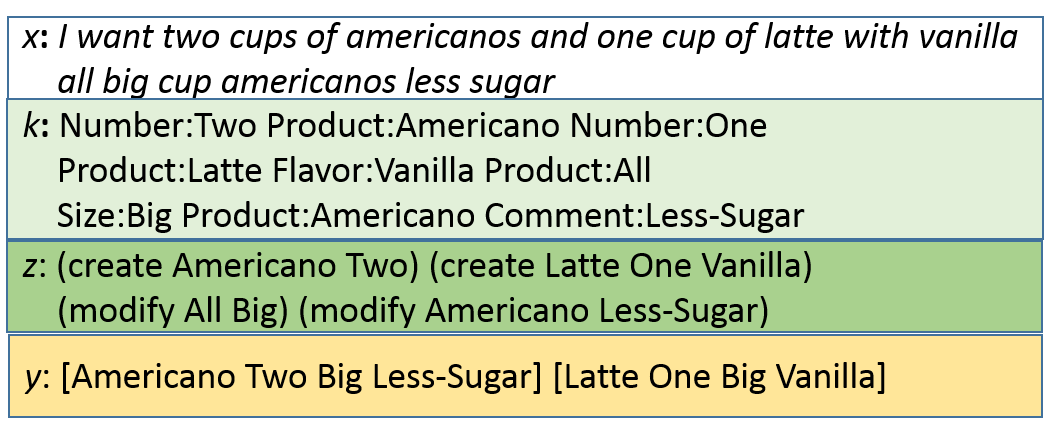}
\caption{Overview of our turn-level SLU setup. Given a turn of utterances \(x\), we first extract the tag sequence \(k\), then parse \(k\) to a program \(z\) that after execution results in the denotation \(y\).}
\centering
\label{fig:example}
\end{figure}

Spoken language understanding (SLU) is a core component of voice interaction applications. Traditionally, SLU is performed on sentences generated by voice activity detection on user queries. In this work, we focus on \textit{turn-level spoken language understanding}, that is, to understand the ultimate intent when a user speaks one or more utterances compositionally in one turn. Figure~\ref{fig:example} illustrates a voice ordering example for turn-level SLU. The user speaks 4 utterances in a sequence to the agent as ``I want two cups of americanos and one cup of latte with vanilla all big cup americanos less sugar". The agent interprets the utterances as two order creation actions and two order modification actions, and finally executes the actions and generates the order automatically as ``two big cups of americanos with less sugar and one big cup of latte with vanilla".

Turn-level SLU consists of multiple sub-tasks. Firstly, spoken language contains disfluencies (e.g. repetitions or repairs) and no explicit structures (e.g. sentence boundaries). Hence, disfluency removal and sentence segmentation are necessary for the downstream language understanding component. Secondly, coreference resolution, intent segmentation and classification, and slot extraction are needed for inferring the ultimate intent of multiple utterances. As shown in Figure 1, the first mention of ``americanos" refers to the product ``americanos" in an order creation action. The second mention of ``americanos", through coreference resolution, is decided as referring to the previously mentioned ``americanos", and triggers an order modification action. 

A traditional pipelined approach solves the aforementioned sub-tasks through a sequence of components. However, development of these components usually requires non-trivial annotation effort for supervised training and it is not easy to port the pipeline to new domains or scale it up. To address these problems, we propose an end-to-end statistical model with weak supervision. Weakly supervised learning has been extensively studied in the fields of semantic parsing and program synthesis~\cite{Cheng:17,Guu:17,Krishnamurthy:17,Liang:17,Rabinovich:17, Suhr:18a,Suhr:18b,Goldman:18,Liang:18}, where indirect supervisions (e.g. question-denotation pairs) are adopted. Direct supervisions (e.g., question-program pairs) require annotating programs, but annotating programs is known to be expensive and difficult to scale up. Compared to direct supervisions, training data for indirect supervisions are easy to collect with low cost. The end-to-end learning approach with weak supervision can scale up easily compared to supervised learning approaches.

It is challenging to develop a semantic parser for turn-level SLU based on question-denotation pairs.
\textit{Firstly}, there is a large search space for
program exploration. As shown in Figure~\ref{fig:example},
a user speaks multiple utterances in a session, where a
utterance can have different intents (e.g. creation or
modification) and slots. Incorrect interpretations of utterances, and incorrect programs \(z\), can accidentally produce the correct target denotations. These incorrect programs are denoted \textit{spurious programs}. Sophisticated algorithms are
required for solving complicated problems with long 
promising trajectories and guarding against spurious programs. \textit{Secondly}, the complexity
of the problems has high divergence. Some turns only have
one utterance (easy task), while others
may have much more utterances (hard task).  Policy gradient estimates for long trajectories tend to
have more variance with weak supervision~\cite{Liang:18}.
The hard task needs more diverse and massive amounts of training data than the easy task. As a result, uniformly sampling data for exploration and training suffers from sample inefficiency and over-fitting problems.


In this work, we propose randomized beam search with
memory augmentation (RBSMA) for improving exploration of long and
promising programs for complicated problems. Randomized beam search can improve exploration efficiency~\cite{Guu:17}.
With the enhancement of memory, RBSMA can learn from failed trials 
and guide the
exploration towards unexplored promising directions. With the cache of highest reward programs per turn, RBSMA can re-sample highest reward programs despite the adopted randomized exploration strategy.


Curriculum learning (CL) can deal with the diversity of sample complexity~\cite{Liang:17}, where the learner focuses on easy ones at first, then gradually puts more weights on more difficult ones. Despite great empirical results, most CL methods are based on hand-crafted curricula. In this way, an expert defines the level of
complexity of a sample and designs curricula, which makes the approach difficult to scale up for complicated problems. In this work, we extend automated CL approaches for supervised learning ~\cite{Graves:17} to weakly supervised learning. We use self reward gain as a signal of reward, which measures the learning progress when the learner is fed with a batch of
question-denotation pairs sampled from some tasks.
Then, the reward is used in a nonstationary multi-armed bandit setting, which then determines a stochastic syllabus and provides training
data to the learner in a principled order.


We evaluate our proposed model for turn-level SLU using real-world user logs of a commercial voice ordering system. Experimental results demonstrate that when trained on a small number of end-to-end annotated sessions collected with low cost, the proposed model performs comparably to the deployed pipelined system, saving the development labor over an order of magnitude. 
In particular, RBSMA improves the test set accuracy by 7.8\% relative compared to the standard beam search. Automated CL leads to better generalization and further improves the test set accuracy by 5.0\% relative, reaching an overall 10.4\% relative gain over standard beam search without CL.

It should be noted that the proposed model is not limited to the voice ordering applications. It can be readily applied in other voice interaction systems for turn-level SLU (e.g., constrain search space by understanding user's multiple queries compositionally), semantic parsing and program synthesis, among others. The technical contributions in this work are as follows:
\begin{enumerate}
\item We develop an end-to-end statistical model with weak supervision (denotations) for turn-level SLU and find that it can perform well, easily scale up and port to new domains.
\item We propose randomized beam search with memory augmentation (RBSMA). We show that RBSMA can explore long promising trajectories for complicated problems more efficiently than the standard beam search.
\item We develop an automated curriculum learning approach for weakly supervised learning to address the diversity of problem complexities.  We  observe that automated CL can lead to faster training and better generalization.
\end{enumerate}

\section{Related Work}
\label{sec:related}
Recently there has been a lot of progress in learning neural semantic parsers with weak supervision ~\cite{Cheng:17,Guu:17,Krishnamurthy:17,Liang:17,Rabinovich:17, Suhr:18a,Suhr:18b,Goldman:18,Liang:18}.  
Systematic search was explored to improve exploration of reinforcement learning (RL) and stability of 
 weak supervision~\cite{Guu:17,Liang:17}. Memory Augmented Policy Optimization (MAPO)~\cite{Liang:18} was proposed using a memory buffer of promising trajectories to reduce the variance of policy optimization. In this work, we extend these ideas by proposing randomized beam search with memory augmentation to improve the exploration efficiency.
CL can deal with the diversity of sample
complexity~\cite{Liang:17}. In this work, we develop an effective approach extending automated CL for supervised learning~\cite{Graves:17} to CL for weakly supervised learning.


\section{Task Description}
\label{sect:task}
\subsection{Two General Tasks}
\label{subsect:twotasks}
Inspired by~\cite{Goldman:18}, turn-level SLU is roughly divided into a \textit{lexical} task (i.e., mapping words and phrases to tags that are parts of a program) and a \textit{structural} task (i.e., combining tags into a program).
We collect a typed dictionary and their aliases, and conduct the lexical task by word matching.
Figure~\ref{fig:example} shows the matched tag sequence as ``Number:Two Product:Americano Number:One Product:Latte Flavor:Vanilla Product:All Size:Big Product:Americano Comment:Less-Sugar". 
The tagging process filters out noise and task-irrelevant words in the automatic speech recognition (ASR) output, easing the downstream structural task.

The structural task aims at generating the target action sequence (program) based on the tag sequence. Each action in the sequence is a token of the program. The action sequence is then executed to generate the final denotation. The target action sequence for the above tag sequence is ``(create Americano Two) (create Latte One Vanilla) (modify All Big) (modify Americano Less-Sugar)". In our turn-level SLU setup, the structural task is quite challenging due to multiple types of tag manipulations: (1) \textit{Tag Deletion}: Repetitive tags created due to disfluency should be removed, for example, ``americano americano big cup" should be transformed to ``(create Americano Big)". (2) \textit{Tag Segmentation}: Intent segmentation will group a set of tags and generate a corresponding action. For example, ``two big americano cold one latte" should be segmented into two actions, ``(create Americano Two Big Cold) (create Latte One)"\footnote{We use the heuristics of segmentation based on the Number tag, hence ``cold" is grouped into the first action.}. (3) \textit{Tag Copy and Assignment}: For nested structures, tags on the root node should be copied and assigned to leaf nodes. For example, ``two hot lattes \textit{one big cup one small cup}" should be transformed to ``(create Latte One Big Hot) (create Latte One Small Hot)" . (4)
\textit{Tag Global Assignment}: Some tags should be assigned to the node with a long distance for the modification purpose, that is, co-reference resolution is implicitly modeled. For example, ``one americano two lattes americano big cup" should be interpreted as ``(create Americano One) (create Latte Two) (modify Americano Big)".

\subsection{Problem Statement}
Given a training set of \(N\) examples \(\{(x_i, k_i, y_i)\}_{i=1}^{N}\), where \(x_i\) is a sequence of utterances within a turn, \(k_i\) is the tag sequence of \(x_i\) that is produced by the \textit{lexical} task, \(y_i\) is a set of objects that the agent should generate according to \(x_i\).
Our goal is to learn a semantic parser that maps a turn of utterances \(x\) to a program \(z\), such that when \(z\) is executed by the agent, it yields the correct denotation
\(y\).

\subsection{Program}
In our turn-level SLU setup based on voice ordering, the objects in the denotation \(y\) have internal structures. For example, an ordered object refers to one product, which contains several properties such as product name and number of cups (Section~\ref{sect:dataset}). Based on studying real-world user logs of a commercial voice ordering system, we use two kinds of functions 
for the tokens in a program: (1) \textit{Create Function}: (create \(p_1 \cdots p_m\)) (2) \textit{Modify Function}: (modify \(p_1 \cdots p_m\)). Here \(p_1\) is the key property (e.g. product name) and is mandatory, while the other properties are optional and will be set to default values when missing (e.g. \(p_2\) refers to number of cup, its default value is one), \(m\) denotes the number of properties, and \(p_1 \cdots p_m\) are properties and also parameters for a function. 

\section{Model Description}
We decompose the program generation task into two subtasks. One is function type (function name) generation, depending on the context; the other is selection of the set of tags produced by the lexical task as parameters for each function type. We utilize a seq2seq model based on the semantic parser in~\cite{Guu:17,Goldman:18} and extend it with pointer-generator~\cite{See:17}. In this way, the decoder vocabulary size is significantly reduced. 



The probability of a program is the product of the probabilities of its tokens given the history: \( p_\theta(z|x) \approx p_\theta(z|k) =  \prod_{t} p_\theta(z_t|k, z_{1:t-1}) \). We approximate the conditional probability of a program \(z\) given the input turn \(x\) by the tag sequence \(k\) given \(x\). 
\(p_\theta(z_t|k, z_{1:t-1})\) is computed as \(p_{gen,t}p_{vocab\_func}(z_t|k, z_{1:t-1}) 
 + (1-p_{gen,t})\sum_{i:z_t=k_i}{\alpha_{t,i}}\), where \(p_{gen,t}\) is probability of the function name generation subtask (rather than selection) for timestep \(t\), \(p_{vocab\_func}(z_t|k, z_{1:t-1})\) is probability of generating function name \(z_t\), \(\alpha_{t,i}\) is the attention weight.

\section{Learning}
\label{sect:learning}
We now describe our exploration-based learning algorithm with weak supervision. To use weak supervision, we treat the program \(z\) as a latent variable that is approximately marginalized. To describe the learning objective, we define \(R(z, y)=\mathbbm{1}(y^{'}=y)-||y^{'}-y||\), where \(y^{'}\) is the execution result of \(z\), the first part \(\mathbbm{1}(y^{'}=y)\in\{0,1\}\) is a binary signal indicating whether the task has been completed by producing the target denotation \(y\), the second part \(-||y^{'}-y||\) computes the edit distance
between the execution result from \(z\) and the target denotation, providing a meaningful signal for uncompleted task situations.

The objective is to maximize the following function:
\begin{equation*} \label{eq2}
\begin{split}
\sum_{z\in Z}p_\theta(z|x)R(z, y) \approx \sum_{z\in B}p_\theta(z|x)R(z, y)
\end{split}
\end{equation*}
where \(Z\) is the program space, and \(B \subset Z\) are the programs found by beam search.

\subsection{RBSMA : Randomized Beam Search with Memory Augmentation}
Beam search is a powerful approach for facilitating systematic search through the large space of programs for training with weak supervision. Typically, at each decoding step, we maintain a beam \(B\) of program prefixes of length \(n\), then expand the program prefixes fully to program pool \(P\) of
length \(n+1\) and keep the top \(|B|\) program prefixes with the highest model probabilities out of \(P\).

We explore randomized beam search~\cite{Guu:17}, which combines the standard beam search with the randomized off-policy exploration of RL. Extensive studies in RL show that noise injection in the action space (i.e., when decoding program tokens) 
can significantly improve the exploration efficiency. For weak supervision, randomized beam search can increase the chance of finding correct programs. Instead of keeping the top \(|B|\) scored program prefixes at each decoding step, we either uniformly sample a program prefix out of \(P\) with probability \(\epsilon\) or pick the highest scoring program prefix
in \(P\) with probability \(1-\epsilon\) (\(\epsilon\)-greedy method).
However, for turn-level SLU, the program space is very large. Although this randomized strategy can aid exploration, it could repeatedly sample the same incorrect programs over time, since most programs in the program space are incorrect. Hence exploration is still guided by the current model policy and long tail promising trajectories are difficult to be explored. For example, most turns only contain creation actions, therefore the model will assign high probabilities for creation actions. Thus, it is difficult to explore target programs that are composed of both creation and modification actions.

To address these problems, we extend randomized beam search with memory augmentation, i.e., RBSMA. We maintain a set of fully explored program prefixes \(C^{e}\) for turn \(x\). We first filter out fully explored program prefixes in \(P\) that exist in \(C^{e}\), then select the \(|B|\) programs in the remaining \(P\) using the \(\epsilon\)-greedy method. In a sense, \(C^{e}\) enables us to learn from failed trials, considering that most prefixes in \(C^{e}\) refer to incorrect programs. Hence this approach helps guide the exploration towards unexplored promising directions. However, even if we can sample the correct program for once, we can hardly re-sample it due to the adopted randomized exploration strategy, which makes the learning process difficult to converge. Therefore, we maintain a cache of highest reward programs \(C^{p}\) explored so far for each turn \(x\). After a procedure of beam search, we augment the beam search result \(S\) with programs in \(C^{p}\) and update the highest reward programs in \(C^{p}\) with \(S\). The pseudo code for RBSMA is shown in Algorithm~\ref{alg:rbsma}.

\begin{algorithm}
\caption{RBSMA}
\hspace*{0.02in} {\bf Input:}
turn \(x\), fully explored program prefixes \(C^{e}\), cache of highest reward programs \(C^{p}\), number of decoding steps as \(T\), program pool \(P\), beam \(B\) of size \(|B|\), program \(z_{1:t}\) of length t\\
\hspace*{0.02in} {\bf Output:}
beam search result \(S\)
\begin{algorithmic}[1]
\State \(B_1 \gets\)  compute beam of programs of length 1
\For{\(t\) = \(2\)...\(T\)}
\State empty \(P\)
\For{\(s \in B_{t-1}\)} \Comment{Decode}
\State \(s\)=\(z_{1:t-1}\)
\State \(\#\) cont(\(s\)): output from one decoding step
\State cont(\(s\))=\(\{z_{1:t}|z_{1:t-1}, z_{1:t}、\notin C^{e}\}\)
\If{cont(\(s\)) is empty}
\State \(\#\) s:fully explored program prefix
\State insert s in \(C^{e}\)
\EndIf
\State \(P\).add(cont(\(s\)))
\EndFor
\State \(B_t \gets\) \(|B|\) programs from \(P\) with \(\epsilon\)-greedy
\EndFor
\State \(S=B_T \cup C^{p}\)
\State \(C^{p}\).update(\(S\))
\State \Return \(S\)
\end{algorithmic}
\label{alg:rbsma}
\end{algorithm}

Recently MAPO~\cite{Liang:18} was proposed as using a memory buffer of promising trajectories to reduce the variance of policy optimization for program synthesis. There are two major differences between our proposed RBSMA and MAPO. Firstly, RBSMA is based on beam search and MAPO employs Monte Carlo (MC) style sampling. The MC style sampling methods tend to revisit the programs with the highest distribution; whereas, after the highest probability program in a peaky distribution under the model policy, beam search still can use its remaining beam capacity to explore at least \(B-1\) other programs.
Secondly, RBSMA utilizes a randomized exploration strategy, which has been proved to improve the efficiency of exploration and is critical for solving the complicated problems in turn-level SLU.   

\subsection{Automated Curriculum Learning}
\label{sect:autocl}

Considering the diversity of problem complexity,
and to facilitate faster and better learning, we explore automated CL that organizes data into a curriculum and presents it in a principled order to the learning algorithm. We organize the training set \(\{(x_i, k_i, y_i)\}_{i=1}^{N}\) into \(M\) tasks \(\{D_i\}_{i=1}^{M}\).
An ensemble of all the tasks  \(\{D_i\}_{i=1}^{M}\) is a curriculum. A sample \(b\) is a batch of data \(\{(x_i, k_i, y_i)\}_{i=1}^{|b|}\) drawn randomly from one of the tasks.

Inspired by~\cite{Graves:17}, we view a curriculum containing \(M\) tasks as an \(M\)-armed
bandit and design a syllabus as an adaptive policy which seeks to
maximize payoffs from this bandit and continuously adapts to optimize the learning progress. An agent selects a sequence
of arms (tasks) \(D_1,..,D_T\) over T rounds. After each round, the
selected task produces a payoff (real-valued reward) \(r_t\) and the
payoffs for the other tasks are not observed.

The bandit is non-stationary because the parameters related to \( p_\theta(z|x)\) update during training. Therefore, the payoff for each arm (task) can change between successive choices. Following~\cite{Graves:17}, we use adversarial bandits, denoted \textit{Exp3.S} algorithm~\cite{Auer:02,Graves:17}, as shown below:
\begin{equation*} \label{eq3}
\begin{split}
\pi_t(i) &=  (1-\epsilon)\frac{e^{w_{t,i}}}{\sum_{j=1}^M e^{w_{t,j}}}+\frac{\epsilon}{M} \\ 
w_{t,i} &= log[(1-t^{-1})exp\{w_{t-1,i}+\eta \tilde{r}_{t-1,i}\}           \\
& +\frac{t^{-1}}{M-1}\sum_{j\ne i}exp\{w_{t-1,j}+\eta \tilde{r}_{t-1,j}\}]  \\
\tilde{r}_{s,i} &= \frac{r_s\mathbbm{1}(a_s=i)}{\pi_s(i)}
\end{split}
\end{equation*}
where \(\pi_t\) is policy that is defined by a set of weights \(w_{t,i}\), \(\epsilon\) refers to the extent of noise injection, \(r_s\) is the observed reward at round \(s\),\(a_s\) is the arm selected at round \(s\) from \(\pi_t\) based on estimated bandit probability distributions of success, \(\tilde{r}_{s,i}\) is re-scaled reward for arm \(i\), \(\eta\) is the learning step.

Different from the loss-driven progress signals explored in~\cite{Graves:17} for supervised learning, for weak supervision, we consider self reward gain (SRG) as the learning progress signal, by comparing the predictions made by the model before
and after training on some sample \(b\). We denote the model
parameters before and after training on \(b\) by \(\theta\) and \(\theta^{'}\), respectively. To avoid bias, we sample another \(b^{'}\) from the same task of \(b\). 
\[SRG=\hat{R}(b^{'}, \theta^{'})-\hat{R}(b^{'}, \theta)\]
where program \(z_\theta\) is predicted by model \(\theta\) on \(x\) out of \(b^{'}\); \(\hat{R}(b^{'}, \theta)\) equals \(R(z_\theta,y)\) that was defined early in Section~\ref{sect:learning}; \(\hat{R}(b^{'}, \theta^{'})\) is computed similarly using model \(\theta^{'}\).

Finally, we re-scale \(SRG\) to the interval of \([-1,1]\) by min-max normalization for better convergence and assign the rescaled SRG to payoff \(r_t\). The pseudo code for Automated CL with SRG is shown in Algorithm~\ref{alg:autocl}. 
\begin{algorithm}
\caption{Automated Curriculum Learning with SRG}
\hspace*{0.02in} {\bf Initially:}
\(w_{1,i}=0\) \\
\begin{algorithmic}[1]
\For{\(t\) = \(1\)...\(T\)}
\State \(\pi_t(k) =  (1-\epsilon)\frac{e^{w_{t,k}}}{\sum_{j=1}^M e^{w_{t,j}}}+\frac{\epsilon}{M}\)
\State Draw task index \(k\) from \(\pi_t\)
\State Draw training sample \(b\) from \(D_k\)
\State Train network \(p_{\theta}\) on \(b\), result in \(p_{\theta^{'}}\)
\State Draw another sample \(b^{'}\) from \(D_k\)
\State \(SRG=\hat{R}(b^{'}, \theta^{'})-\hat{R}(b^{'}, \theta)\)
\State Map \(SRG\) to \(r_t \in [-1,1]\)
\State Update \(w\) with reward \(r_t\) using Exp3.S
\EndFor
\end{algorithmic}
\label{alg:autocl}
\end{algorithm}

\section{Experiments and Analysis}
\label{sect:expts}

\subsection{Dataset}
\label{sect:dataset}
The example application is voice ordering for coffee. 
One item (object) in a coffee order contains seven properties, summarized in Table~\ref{tab:properties}. For evaluation, we only need to collect question-denotation pairs, that is, a session (turn) of user utterances and its final order. We find that the final order is easy to annotate and the weak supervision data can be collected with low cost.

\begin{table}[h!]
\centering
\begin{tabular}{|c| c| c| c|} 
 \hline
 Property Name & Examples & Total \# \\ [0.5ex] 
 \hline\hline
 product & americano, latte.. & 16  \\ 
 \hline
 number & one, two, three.. & 20  \\
 \hline
 cup size & small, middle,big & 3  \\
 \hline
 flavor & vanilla, caramel.. & 10  \\
 \hline
 hot & cold, hot & 2  \\ 
 \hline
 location & pack, dine in & 2  \\
 \hline
 comment & less sugar, little ice.. & 18  \\
 \hline
\end{tabular}
\caption{Properties of an ordered item.}
\label{tab:properties}
\end{table}

We create two datasets, namely, \textit{recorded100} and
\textit{log1144}. \textit{recorded100} is composed of 100 sessions by
recording customers making orders by talking to a human clerk in a coffee shop,
with the orders generated manually by the clerk. 
\textit{log1144}
is composed of 
1144
sessions extracted from real-world user logs of a commercial voice ordering system in a coffee shop, where the orders are labelled manually. After manually transcribing user utterances based on ASR output, \textit{recorded100} is used as the training set and
\textit{log1144} as the test set. 
The data statistics are summarized in Table~\ref{tab:stats}. One major goal of the proposed model is to use a small amount of weak supervision data collected with low cost to train a high-performing turn-level SLU system. Hence we intentionally train on a small amount of training data (\textit{recorded100}) and test on a much larger test set to evaluate generalization of the proposed model. 

\begin{table}[htb]
\centering
\begin{tabular}{|c| c| c| c| c|} 
 \hline
 Dataset & \#Sessions & r1 & r2 & r3 \\ [0.5ex] 
 \hline\hline
 \textit{recorded100} & 100 & 47\%  & 45\% & 8\%\\ 
 \hline
 \textit{log1144} & 1144 & 62\%  & 29\% & 9\% \\
 \hline
\end{tabular}
\caption{Statistics of the training set \textit{recorded100} and test set \textit{log1144}.}
\label{tab:stats}
\end{table}

In both datasets, we observe users order up to three different items in a session. Columns r1, r2, and r3 in Table~\ref{tab:stats} show the percentage of sessions with ordering one, two, or three items, respectively~\footnote{An example of ordering three items (r3) in the test set is ``one middle cup of mocha and one big cup of latte with vanilla two cups of regular lattes all take away".}. Sessions with one ordered item (easy task) are much more than sessions with three ordered items (the most complex problems in this setup). 
Note that one item has seven different properties, and both creation and modification actions might be included in the session. Hence, the program space for r3 is extremely large. We evaluate our proposed model on the training and test sets. The evaluation metric is accuracy, i.e., the percentage of the execution result \(y^{'}\) based on the generated program \(z\) equaling the target result \(y\).



\subsection{The Pipelined Baseline}
We take the deployed pipelined system as the baseline. In this system, first, we transform the utterances in a turn into a sequence of tags based on the \textit{lexical} task (Section~\ref{subsect:twotasks}). Second, we remove contiguous repeated tags for disfluency removal. Third, inspired by the shift-reduce parser, we maintain a stack of tags and a set of ordered items. Initially, we empty the stack and the set. Then, we look ahead each of un-scanned tags, and make decisions of actions based on hand-crafted rules. Some decision may shift the current tag to the stack, while others reduce the current stack. Then we decide whether the reduce action is a creation action or a modification action also based on hand-crafted rules. 
The baseline approach scales up poorly (e.g., for more combinations of order items) and is difficult to port to new domains.

\subsection{Details of Model and Training}
The seq2seq model for program generation consists of a BiLSTM encoder and a feedforward decoder with dimensions of hidden states as 30 and 50, respectively.
The decoder takes input encoder hidden states as well as embeddings of the last 5 decoded tokens and bag-of-words vector of all the decoded tokens. The decoding beam size is 40. Token embedding dimension is 12. Encoder input tag embeddings are initialized as follows. Given \(n\) types of tags in total, the tag embedding dimension is \(1+n*2\). The first dimension is the index of this tag in the set of all tags; the \((2i-1)^{th} (i \in [1,n])\) dimension has value 1 or 0, indicating whether the tag is in type \(i\) or not; the \(2i^{th}\) dimension is the index of the tag in type \(i\). Tag embeddings are then optimized end-to-end.  

We define \(||y^{'}-y||\) in \(R(z, y)\) as the total number of
different properties between items in \(y^{'}\) and in \(y\).  To train the parameters
\(p_\theta(z|x)\), 
similar to~\cite{Guo:18},
we re-scale \(R(z, y)\) based on its
ranking for better convergence and optimization towards
better programs. We set the
reward of programs with top-ranked \(R(z, y)\) as 1.0, and
set the reward for the rest as 0. Note that the original \(R(z, y)\) is
discrete, thus there are multiple top-ranked programs with
re-scaled reward 1.0. We also employ code assistance to help prune the search space by checking syntax of partially generated programs, following previous weak supervision work~\cite{Liang:17}.  To encourage
exploration with
\(\epsilon\)-greedy, we set \(\epsilon=0.5\) in RBSMA. For automated CL, we define tasks based on the difficulty of the denotation \(y\). A straightforward measure of difficulty is the number of ordered items in \(y\). Here we assign the training data that includes just one ordered item in the denotation to task \(D_1\) (easy task), and data with multiple ordered items to task \(D_2\) (hard task). 
The parameters for the Exp3.S algorithm (Section~\ref{sect:autocl}) are \(\eta=0.1\), \(\epsilon=0.05\). We uniformly sample from all the tasks
for the first 80 steps for warm up. Adam is used for optimization, with learning rate 0.001, and mini-batch 
size of 8\footnote{Hyperparameters are optimized based on the training set accuracy.}.

\subsection{Results}
\label{subsect:results}
We compare the proposed model (denoted WeakSup) to the deployed pipelined system (baseline). For ablation analysis, we evaluate the following variants of our approach. \textit{WeakSup\_SBS\_Uni} is variant of WeakSup based on the standard beam search, i.e., removing the memory in RBSMA and set \(\epsilon=0\), also with uniformly sampling from both easy and hard tasks, i.e., no CL. \textit{WeakSup\_SBS} is variant of WeakSup based on the standard beam search and with CL. \textit{WeakSup\_Uni} is variant of WeakSup with RBSMA but no CL.

\begin{table}[h!]
\centering
\begin{tabular}{|c|c|c|} 
 \hline
 Model & Training & Test \\ \hline \hline
 Pipelined & 95\% & 85.5\% \\ 
 \hline
 WeakSup\_SBS\_Uni & 86\% & 78.1\% \\
 WeakSup\_SBS & 86\% & 80.0\%  \\
 WeakSup\_Uni & 99\% & 82.1\% \\
 WeakSup & 100\% &  86.2\%  \\
 \hline
\end{tabular}
\caption{Accuracy on the training set and test set.}
\label{tab:accuracy}
\end{table}


As shown in Table~\ref{tab:accuracy}, WeakSup performs slightly better than the deployed Pipelined system on the test
set, while trained only on a small number of end-to-end annotated sessions
collected with low cost. We observe that 92.3\% errors made by WeakSup\_SBS on the training set belong to the hard task (i.e., ordering more than one item) since they are more difficult for exploration. WeakSup achieves 100\% accuracy on the training dataset, demonstrating that long promising programs for complicated problems can be explored by RBSMA. Pipelined achieves 95\% accuracy on the training set, indicating that it is difficult to cover all the hard problems through hand-crafted rules. Without CL, RBSMA improves the test set accuracy by 5.1\% relative (from 78.1\% to 82.1\%) over the standard beam search; with CL, 7.8\% relative (from 80.0\% to 86.2\%). 
CL improves generalization on the test data. With RBSMA,  CL achieves 5\% relative gain (comparing WeakSup with
WeakSup\_Uni), while maintaining similar training set accuracy.
The combination of RBSMA and CL improves the test set accuracy by 10.4\% relative (78.1\% to 86.2\%).
The total training time for the proposed model WeakSup is 1.5 hours on one Tesla M40 GPU. Summing up data preparation and computation time, the proposed model saves development effort over an order of magnitude compared to the deployed pipelined system.

\subsection{Analysis of Exploration Strategy}
\begin{figure}[h]
\includegraphics[scale=0.6]{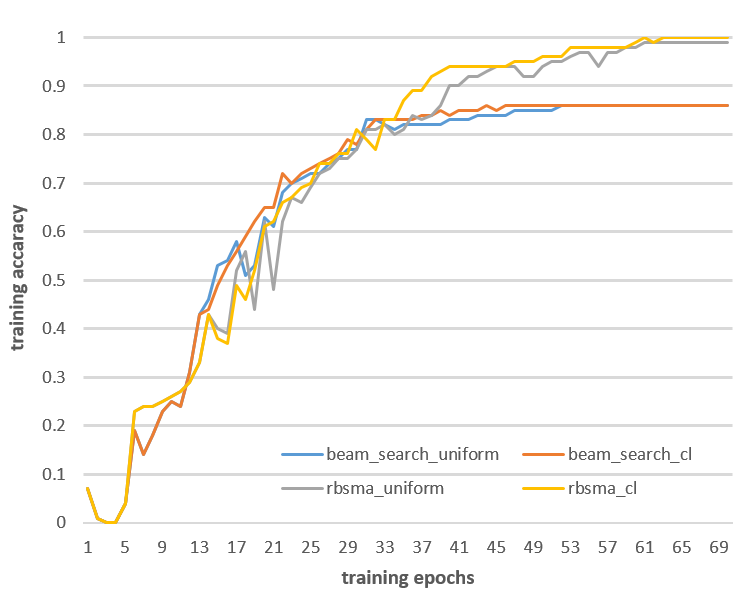}
\caption{Analysis of exploration strategy.}
\centering
\label{fig:exploration2}
\end{figure}

We study exploration strategies comparing the standard beam search vs. RBSMA, automated CL (cl) vs. uniform sampling (uniform). We evaluate beam\_searh\_uniform,
beam\_searh\_cl, rbsma\_uniform, rbsma\_cl. We measure the exploration progress by the training set accuracy for a given epoch, shown in Figure~\ref{fig:exploration2}.

We have two observations. First, exploration with RBSMA progresses slowly at the very beginning compared to beam search (probably due to the randomized exploration strategy), but catches up and solves all the training set problems in the end. In contrast, exploration with the standard beam search stops progressing after epoch 52. Second, automated CL progresses faster than uniform sampling for most of the time. Particularly after epoch 35, rbsma\_cl significantly improves over rbsma\_uniform, probably due to the adaptive policy (sampling more samples from hard task).

\subsection{Analysis of Adaptive Policy of Automated Curriculum Learning}
\begin{figure}[h]
\includegraphics[scale=0.6]{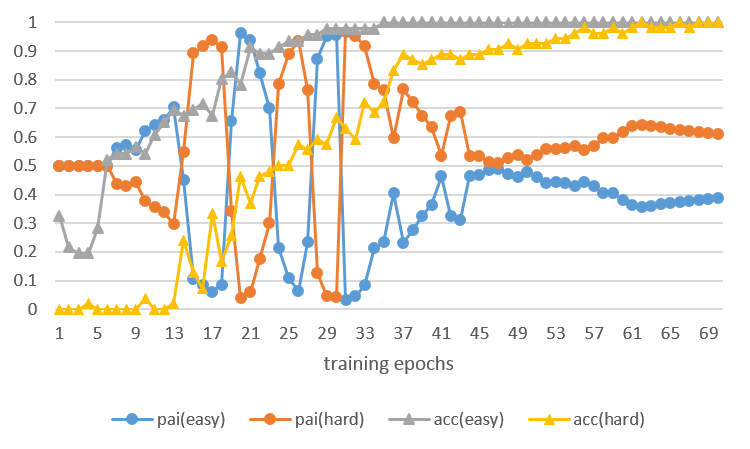}
\caption{Average policy and accuracy over epochs.}
\centering
\label{fig:policy}
\end{figure}

The efficacy of the adaptive policy of our proposed automated CL algorithm is illustrated in Figure~\ref{fig:policy}. Pai(easy) denotes the probability of sampling easy task under the policy, while pai (hard) the probability of sampling hard task. Acc(easy) denotes the accuracy of easy task in the training set, and acc(hard) the accuracy of hard task. Figure~\ref{fig:policy} reveals a consistent strategy, first focusing on easy task, then alternatively on both easy and hard tasks, and finally focusing more on hard task but still sampling from easy task. This automatically learned complex strategy is challenging to achieve even with carefully hand-crafted curricula, due to challenge in defining acceptable performance of tasks. Also, our proposed approach continuously samples from easy task when learning hard ones, effectively addressing the forgetting problem, as acc(easy) in Figure~\ref{fig:policy} does not degrade when acc(hard) improves. In contrast, effectively crafting these mixing strategies is challenging~\cite{Zaremba:14}.

\section{Conclusion}
\label{conclusion}
We present an end-to-end statistical model with weak 
supervision for turn-level SLU. We propose two techniques for better exploration and generalization: (1) RBSMA for complicated problems with long programs, (2) automated CL for weakly supervised learning for dealing with the diversity of problem complexity. Experimental results on real-world user logs show that our model performs comparably to the deployed pipelined system, greatly saving the development labor. Both RBSMA and automated CL significantly improve exploration efficiency and generalization.

\bibliography{acl2019}
\bibliographystyle{acl_natbib}

\end{document}